\documentclass[10pt,twocolumn,letterpaper]{article}

\usepackage[pagenumbers]{iccv}

\usepackage{colortbl}
\usepackage{makecell}
\usepackage{xcolor}
\usepackage{amsmath}  
\usepackage{graphicx}
\usepackage{graphics}
\usepackage{tikz}
\usepackage{siunitx}
\usepackage[misc]{ifsym} 
\newcommand*\circled[1]{\tikz[baseline=(char.base)]{
            \node[shape=circle,fill=black,text=white,draw,inner sep=0.5pt] (char) {#1};}}

\newcommand{\datasetname}{\textit{SynFMC}\xspace}

\newcommand{\methodname}{\textit{FMC}\xspace}
\newcommand{\methodcompletename}{\textit{Free-Form Motion Control}\xspace}

\newcommand{\camctrlname}{CMC\xspace}
\newcommand{\camctrlcompletename}{Camera Motion Controller\xspace}

\newcommand{\objctrlname}{OMC\xspace}
\newcommand{\objctrlcompletename}{Object Motion Controller\xspace}

\definecolor{myorange}{RGB}{255,100,3}
\definecolor{mygray}{gray}{.85}
\definecolor{mygray1}{gray}{.7}
\definecolor{mygray2}{gray}{.93}
\definecolor{mygray3}{gray}{.90}
\usepackage{array}
\newcolumntype{C}[1]{>{\centering\arraybackslash}p{#1}}

\usepackage{pifont}
\newcommand{\xmarkg}{\textcolor{lightgray}{\ding{55}}\xspace}%

\definecolor{iccvblue}{rgb}{0.21,0.49,0.74}
\usepackage[pagebackref=false,breaklinks,colorlinks,allcolors=iccvblue]{hyperref}
\usepackage{svg}

\begin{document}

\title{
\raisebox{-1.5ex}{\protect\includegraphics[height=3.\fontcharht\font`\B]{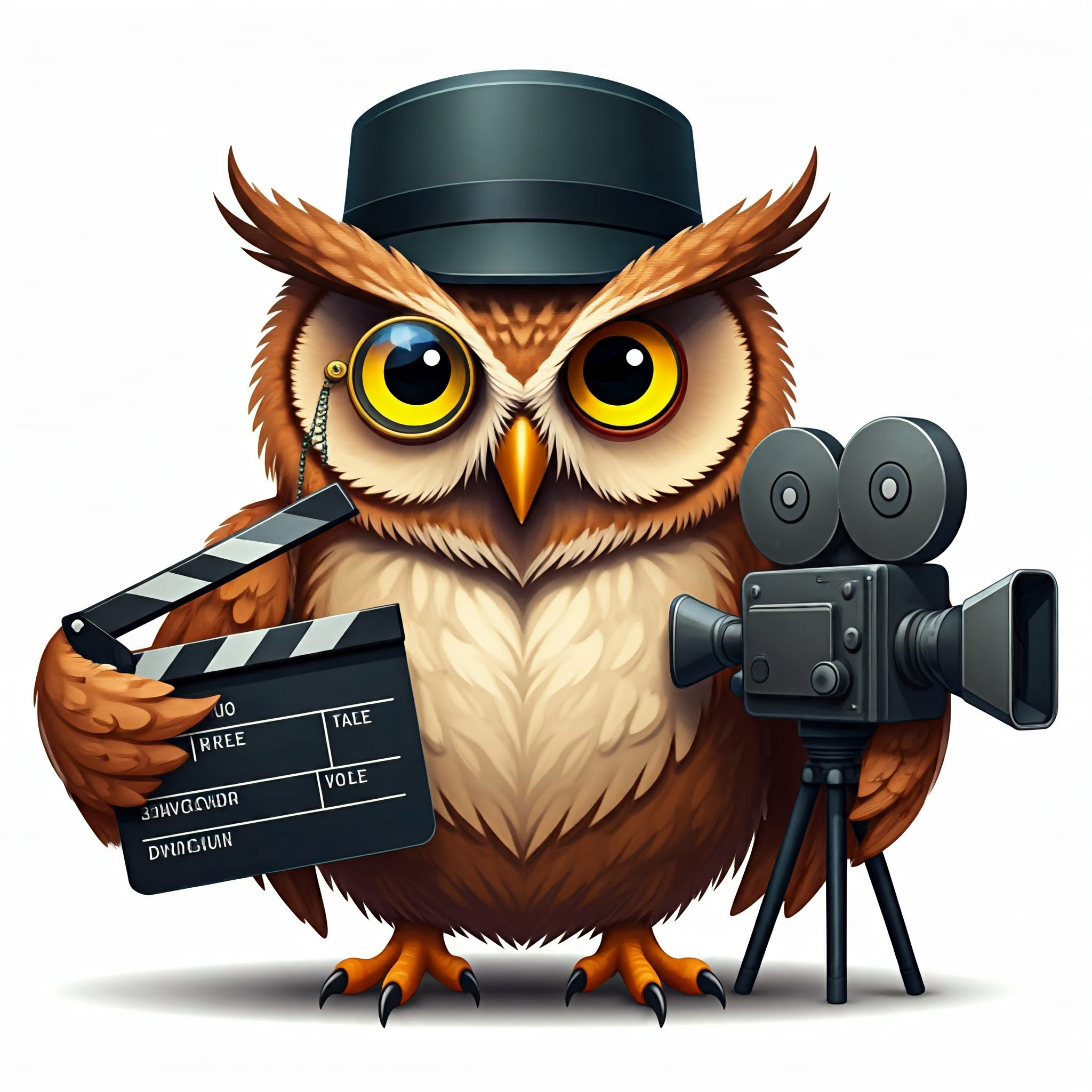}}
{Free-Form Motion Control: Controlling the 6D Poses of Camera \\[-.1em]
and Objects in Video Generation}
}

\author{
Xincheng Shuai$^1$
\quad
Henghui Ding$^1$
\hspace{0.56em}
Zhenyuan Qin$^1$
\hspace{0.56em}
Hao Luo$^{2,3}$
\hspace{0.56em}
Xingjun Ma$^1$
\hspace{0.56em}
Dacheng Tao$^4$
\\
$^1$Fudan University, China
\quad
$^2$DAMO Academy, Alibaba group
\quad
$^3$Hupan Lab
\\
$^4$Nanyang Technological University, Singapore\\
{\tt\footnotesize henghui.ding@gmail.com\quad dacheng.tao@gmail.com}
\\
\href{https://henghuiding.com/SynFMC/}{https://henghuiding.com/SynFMC/}
}

\twocolumn[{%
\renewcommand\twocolumn[1][]{#1}%
\maketitle
\vspace{-7.6mm}
\begin{center} 
\centering 
\captionsetup{type=figure}
\includegraphics[width=0.999\textwidth]{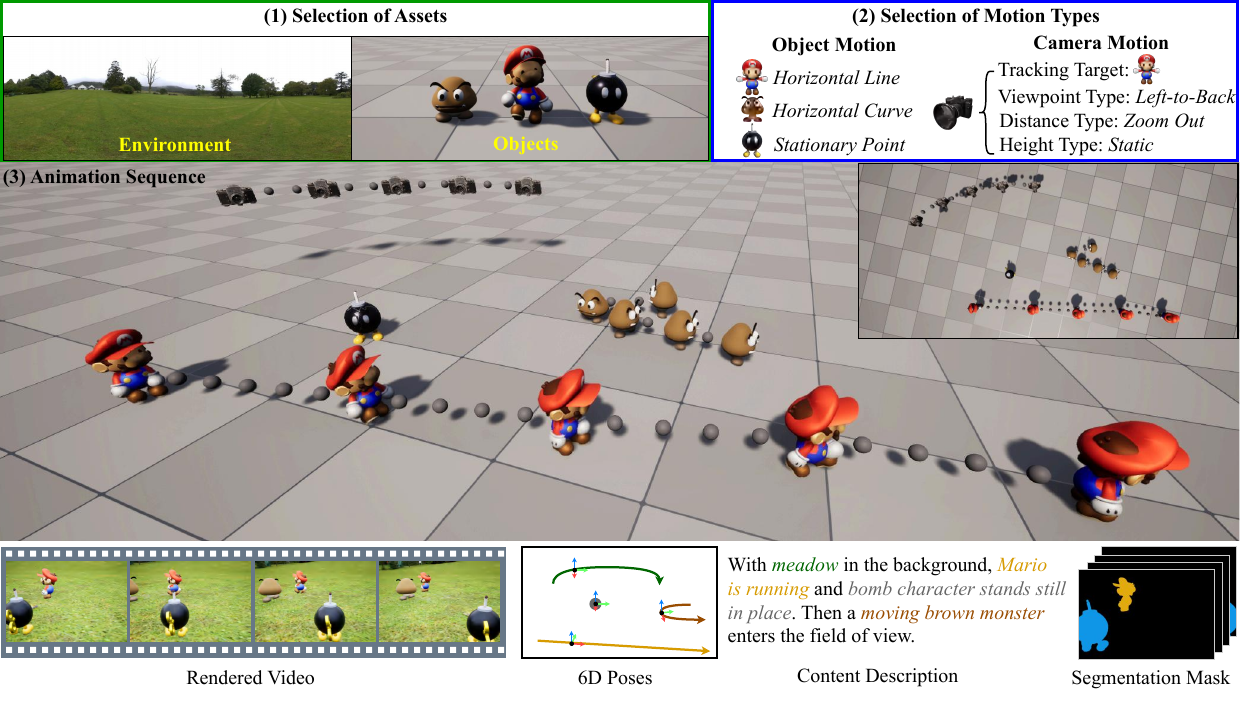}
\vspace{-6.6mm}
\captionof{figure}{The rule-based generation pipeline of videos in the proposed \textbf{\underline{Syn}}thetic Dataset for \textbf{\underline{F}}ree-Form \textbf{\underline{M}}otion \textbf{\underline{C}}ontrol (\textbf{\datasetname}). This example generates synthetic video with three objects: (1) The environment asset and it's matching object assets are selected as the scene elements. (2) The motion types of objects and camera are randomly selected for trajectory generation. (3) The center region shows the resulting 3D animation sequence used for rendering. The rendered video and annotations are demonstrated in the last row.}
\vspace{3.16mm}
\label{fig:teaser}
\end{center}
}]

\maketitle

\begin{abstract}
Controlling the movements of dynamic objects and the camera within generated videos is a meaningful yet challenging task. Due to the lack of datasets with comprehensive 6D pose annotations, existing text-to-video methods can not simultaneously control the motions of both camera and objects in 3D-aware manner, resulting in limited controllability over generated contents. To address this issue and facilitate the research in this field, we introduce a \textbf{\underline{Syn}}thetic Dataset for \textbf{\underline{F}}ree-Form \textbf{\underline{M}}otion \textbf{\underline{C}}ontrol (\textbf{\datasetname}). The proposed \datasetname dataset includes diverse object and environment categories and covers various motion patterns according to specific rules, simulating common and complex real-world scenarios. The complete 6D pose information facilitates models learning to disentangle the motion effects from objects and the camera in a video.~To provide precise 3D-aware motion control, we further propose a method trained on~\datasetname, \methodcompletename (\methodname). \methodname can control the 6D poses of objects and camera independently or simultaneously, producing high-fidelity videos. Moreover, it is compatible with various personalized text-to-image (T2I) models for different content styles. Extensive experiments demonstrate that the proposed \methodname outperforms previous methods across multiple scenarios. 

\end{abstract}

\vspace{-3mm}
\section{Introduction}
\label{sec:intro}
\vspace{-0.6mm}

Controlling motion dynamics in video generation has received increasing attention~\cite{CameraCtrl,MotionCtrl,HumanVid,AnimateDiff,VMC,STDF,MotionClone,DreamMotion,MotionInversion,MotionMaster,shuai2024survey}, as it enables better customization and is crucial in many applications. For example, in filmmaking, directors meticulously choreograph the movements of both actors and the camera.
Consequently, precise control over object and camera motions in video offers creative flexibility.

Despite recent progress, challenges remain in motion control of text-to-video~(T2V) generation. A key limitation is \textit{\textbf{the lack of high-quality datasets with comprehensive 6D pose annotations}}.
For controlling object movement \cite{MotionCtrl,AnyI2V,VideoComposer,DragNUWA,Tora,Trailblazer,FreeTraj,Motion-Zero,MotionBooth,Image-Conductor,Motion-I2V}, the motion is primarily annotated as the trajectory in image space~\cite{MotionCtrl,DragNUWA}. This annotation, however, lacks 3D nature and intertwines the dynamics of both objects and the camera. For example, a rightward trajectory could represent either a stationary camera with a moving object or a static object with a left-moving camera. Recently, \ang{360}-Motion synthetic dataset \cite{3DTrajMaster} provides 6D poses of objects, but limited with static camera setting and motion diversity.
Besides, existing commonly used datasets~\cite{Direct-A-Video,RealEstate-10K,MVImageNet} for learning camera motion mainly focus on scenes with minimal object dynamics.
Some human-centric synthetic datasets \cite{HumanVid,BEDLAM,SynBody,WHAC} provide ground truth for both human subjects and camera motions within a global coordinate system, yet exhibit limited motion diversity and category variety. 

Another limitation is \textbf{\textit{the absence of methods that can independently or jointly control the 6D poses of both object and camera}}. For example, methods like Motion-Zero \cite{Motion-Zero} animate objects without 3D-aware control (\eg, orientation) \cite{PEEKABOO,MotionBooth,Trailblazer}, while methods like CameraCtrl \cite{CameraCtrl} exclusively focus on camera motion~\cite{Training-free-CC,CamCo,VD3D}.~MotionCtrl \cite{MotionCtrl} trains separate modules for object and camera control in a two-stage process.~However, without access to video data containing complete 6D pose annotations for both elements, it struggles to achieve realistic, synchronized control of objects and cameras within the same scene.

\begin{figure}
    \centering
    \includegraphics[width=0.999\linewidth]{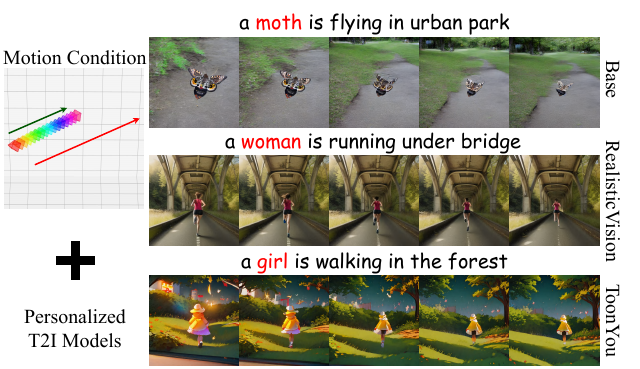}
    \vspace{-7.6mm}
    \caption{Example videos generated by our method \methodname trained on the proposed \datasetname dataset, showing its adaptability  with different personalized T2I models~\cite{StableDiffusion,Toonyou,RealisticVision}.}
    \label{fig:personalized}
    \vspace{-3.6mm}
\end{figure}

To address these limitations, a dataset with comprehensive 6D pose annotations of objects and camera is desired. However, acquiring such data is challenging and typically requires specialized equipment and expertise. In this work, we introduce a \textbf{\underline{Syn}}thetic dataset for \textbf{\underline{F}}ree-Form \textbf{\underline{M}}otion \textbf{\underline{C}}ontrol (\textbf{\datasetname}). Designed with an emphasis on quality and diversity, the dataset includes a rich array of animated object assets and environment assets across various categories. Furthermore, a rule-based generation algorithm is implemented to create trajectories for both objects and the camera, as shown in \cref{fig:teaser}. This algorithm encompasses basic patterns and simulates challenging cinematographic shots as in \cref{fig:curves}. {Compared to recent works~\cite{3DTrajMaster,HumanVid} that can only construct uncontrolled trajectories, our rule-based algorithm supports customized object~\&~camera motions with diverse patterns.} To enhance realism, essential attributes of objects, like living environment, types of speed and size, are annotated using a Multimodal Large Language Model (MLLM)~\cite{InternVL} and manual labeling, facilitating the generation of plausible trajectories. The \datasetname dataset also provides detailed annotations, including 6D pose information of objects and camera, instance segmentation maps, depth maps, and comprehensive descriptions of content and motion, supporting a wide range of research fields.

To further validate the effectiveness of the proposed \datasetname dataset and support {3D-aware control} in T2V generation, we propose a \methodcompletename (\methodname) method. \methodname mainly includes two components: \camctrlcompletename (\camctrlname) and \objctrlcompletename (\objctrlname). Unlike previous methods~\cite{DragNUWA,MotionCtrl}, {Our approach trained on \datasetname disentangles global (camera) and local (object) dynamics and manipulates the 6D poses of camera and objects}. Furthermore, we adopt Domain LoRA~\cite{LoRA} to prevent model from fitting to rendered style in synthetic data. As shown in \cref{fig:personalized} and \cref{fig:lora}, \methodname effectively mitigates the domain gap and adapts to various personalized Text-to-Image (T2I) models, generating high-fidelity results across diverse styles. In addition, \methodname provides flexible user interfaces for motion control. Users can input trajectories for objects and camera by simply drawing 3D curves or by specifying motion types for each (as detailed in \cref{sec:data_pipeline}), which are used by the rule-based algorithm to generate their trajectories accordingly.

\begin{table*}[t]
    \centering
    \renewcommand\arraystretch{1.06}
    \caption{Comparison of the proposed \textbf{\datasetname} with existing datasets. The object/camera motion pattern columns apply only to synthetic datasets. In addition to offering a rich variety of object categories, \datasetname outperforms in motion pattern variety and controllability with comprehensive pose annotations of camera and objects. In our implementation, we only use 26K subset as training data.}
    \vspace{-3.6mm}
    \footnotesize
    \resizebox{1\textwidth}{!}{
    \begin{tabular}{r|ccc>{\centering\arraybackslash}m{0.14\textwidth}>{\centering\arraybackslash}m{0.131\textwidth}>{\centering\arraybackslash}m{0.122\textwidth}>{\centering\arraybackslash}m{0.122\textwidth}}
    \specialrule{.1em}{.05em}{.05em} 
        \rowcolor{mygray!50}Dataset & Clips  & Source &  Category & \makecell[c]{Object Motion\\ Pattern}& \makecell[c]{Camera Motion\\ Pattern}& \makecell[c]{Camera Pose\\Annotation} &  \makecell[c]{Object Motion\\Annotation}  \\ 
        \hline\hline
        RealEstate10K~\cite{RealEstate-10K} & \ \ 65K  & Real & Real Estate &- & -&  Fitting & \xmarkg \\
        \rowcolor{mygray2!36}MVImgNet~\cite{MVImageNet} & 220K  & Real &  \textbf{Common} & - & -&  Fitting & \xmarkg \\

        VideoHD~\cite{DragNUWA} & \ \ 75K & Real & \textbf{Common} & - & - & \xmarkg  & Optical Flow   \\
        
        \rowcolor{mygray2!36}MotionCtrl~\cite{MotionCtrl} & \textbf{240K} &  Real& \textbf{Common} &-  & - & \xmarkg  & Optical Flow \\
        HumanVid-Real~\cite{HumanVid} & \ \ 20K  & Real & Human & - & - & Fitting  & 2D Human Pose \\

        \arrayrulecolor{gray!90}\hline
         BEDLAM~\cite{BEDLAM} & \ \ 10K  & Synthetic & Human & Limited/Uncontrollable & Static & \textbf{Ground Truth}  & 3D Human Pose  \\
        
        \rowcolor{mygray2!36}SynBody~\cite{SynBody} & \ \ 27K  &Synthetic & Human & Limited/Uncontrollable & Static & \textbf{Ground Truth}  & 3D Human Pose \\

        HumanVid-Syn~\cite{HumanVid} & 100K  & Synthetic & Human & Limited/Uncontrollable & \textbf{Diverse}/{Uncontrollable} & \textbf{Ground Truth}  & 3D Human Pose \\

        \rowcolor{mygray2!36}\ang{360}-Motion~\cite{3DTrajMaster} &\ \ 54K  & Synthetic & Animal \& Human & Limited/Uncontrollable & Static & \xmarkg  & \textbf{Object Pose} \\
        
        \arrayrulecolor{black}\hline    
        \rowcolor{green!8}\textbf{\datasetname} (\textbf{ours}) & \ \ 62K  & Synthetic & \textbf{Common} & \textbf{Diverse/Controllable} & \textbf{Diverse/Controllable} & \textbf{Ground Truth}  & \textbf{Object Pose} \\        
     \specialrule{.1em}{.05em}{.05em} 
    \end{tabular}
    }
    \label{tab:datasets}
    \vspace{-5.6mm}
\end{table*}

In summary, our main contributions are as follows:
\begin{itemize}
    \item To the best of our knowledge, the \datasetname dataset is the first to provide 6D pose annotations for both camera and objects. Its diverse scenes and complex motion patterns provide models with valuable resources for learning the dynamics of multiple objects and the camera. 

    \item {The \methodname method can manipulate the 6D poses of the camera and objects independently or simultaneously, achieving high-quality results across diverse scenes.}

    \item Extensive experiments demonstrate that \methodname, trained on \datasetname dataset, generates videos of superior quality compared to state-of-the-art methods.

\end{itemize}

\section{Related Work}
\label{sec:related_work}

\textbf{Dataset with Motion Annotations.}~Most datasets~\cite{MotionCtrl,DragNUWA} focus on operation within the image space.~However, camera and object motions are coupled in this space, while limiting the movement scope. On the other hand, only a few real datasets~\cite{Direct-A-Video,RealEstate-10K,MVImageNet} provide camera pose, and these primarily focus on static scenes without dynamic objects. Some synthetic datasets~\cite{SynBody,BEDLAM,WHAC,HumanVid} provide pose annotations for both objects and camera, yet their 3D assets are predominantly human-centric, limiting category diversity. More discussions are in Sec~\ref{sec:dataset3.1}.

\noindent\textbf{Motion Control Methods.}
Most existing works \cite{CameraCtrl,PEEKABOO,CamCo,Tora,3DTrajMaster} can only control either object motion or viewpoint change. For methods that support manipulation of both the camera and objects, Direct-a-Video~\cite{Direct-A-Video} can only simulate basic movements and simultaneous control in MotionCtrl~\cite{MotionCtrl} often results in suboptimal outcomes as noted in its study~\cite{MotionCtrl}. More discussions are in \cref{sec:approach}.

\section{SynFMC Dataset}
\label{sec:dataset}

\subsection{Comparison with Existing Datasets}
\label{sec:dataset3.1}

There is currently a lack of datasets that contain 6D poses of both objects and the camera~\cite{RealEstate-10K,MVImageNet,HumanVid}. As shown in \cref{tab:datasets}, only a few real datasets~\cite{RealEstate-10K,MVImageNet} provide estimated camera poses and are primarily limited to scenes without dynamic objects due to the suboptimal performance of estimation methods~\cite{COLMAP,DROID-SLAM,ParticleSfM}. Some methods~\cite{DragNUWA,MotionCtrl} use in-the-wild videos with image-space object trajectory inferred by optical flow models~\cite{ParticleSfM}, but this entangles object and camera motions while lacking 3D information like orientation.~Synthetic datasets conveniently obtain pose information~\cite{BEDLAM,WHAC,HumanVid,SynBody}, but most focus on human and are limited to small movements and simple camera motion patterns, limiting the ability to learn complex dynamic.

To facilitate model learning to control the motions in 3D-aware manner, it is essential to construct a new dataset with comprehensive object and camera poses. However, this is highly challenging in real world. First, capturing videos with complex, irregular object and camera motions is extremely difficult, typically requiring specialized equipment and expertise. Then, obtaining accurate pose estimation is difficult.~Devices capable of capturing 6D poses for camera or objects are expensive and difficult to operate. Some studies~\cite{RealEstate-10K,MotionCtrl} attempt to obtain camera pose via estimation models. However, existing algorithms~\cite{ParticleSfM,DROID-SLAM} are time-intensive and often struggle with monocular videos containing dynamic objects. Besides, inferring 6D poses for general objects remains challenging. To address these limitations, we introduce \datasetname, a synthetic dataset generated using \emph{Unreal Engine}~\cite{unrealengine}, containing animations with diverse motion patterns and complete annotations.

\noindent$\bullet$
\textbf{Difference from \ang{360}-Motion Dataset~\cite{3DTrajMaster}.}~\textbf{1)} We handle both static\&dynamic cameras, enabling more complex shots than \cite{3DTrajMaster}'s static setup.~\textbf{2)} Our rule-based algorithm supports diverse (non-)horizontal object motions, unlike \cite{3DTrajMaster}'s GPT-derived horizontal-only ones.~\textbf{3)}~Our~environment \& object assets extend beyond \cite{3DTrajMaster}'s terrestrial scenes.

\begin{figure}
    \centering
    \includegraphics[width=0.999\linewidth]{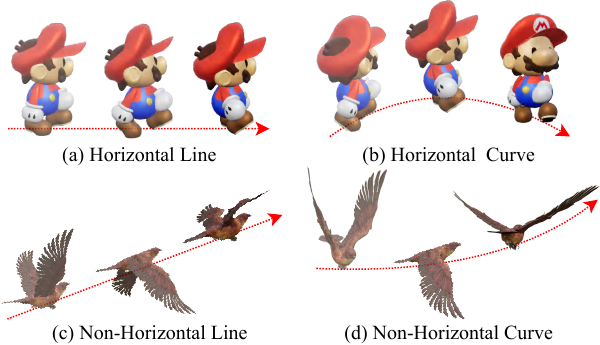}
    \vspace{-6.36mm}
    \caption{{Object motion types. The trajectory of \textit{stationary point} is not presented in the figure, which is a fixed point in the space.}}
    \label{fig:trajectory_shape}
    \vspace{-3.16mm}
\end{figure}

\noindent$\bullet$
\textbf{Difference from HumanVid-Syn Dataset~\cite{HumanVid}.}{~\textbf{1)} The object trajectories of~\cite{HumanVid} rely on predefined 3D motion assets (SMPL-X/skeleton), whereas our rule-based algorithm enables diverse patterns, \eg, in-place/(non-)horizontal motions in \cref{fig:trajectory_shape}.~\textbf{2)}~\cite{HumanVid} randomly samples camera positions within a semi-cylinder in front of human, while we achieve fine-grained control by decoupling camera motion (\cref{fig:camera_type}), supporting controllable and diverse movements. \textbf{3)}~\cite{HumanVid} focuses on synthesizing single-human animation, whereas we support multi-object scenarios, generating richer dynamics.}

\subsection{Overview of \textbf{\datasetname} Dataset}

\datasetname contains 62K videos divided into four groups: 15K \emph{static single-object}, 15K \emph{static multi-object}, 16K \emph{dynamic single-object}, and 16K \emph{dynamic multi-object}. 
\emph{Static} means fixed object locations in world space while the camera remains movable. For diversity, the dataset includes common objects across various categories, such as humans, animals, plants and vehicles, as well as a wide range of environments like streets, grasslands, skies, oceans, \etc. Additionally, \datasetname has diverse and complex multi-object and camera movements, covering not only basic motions but also shots that are challenging to achieve in real-world settings. Object assets are annotated with attributes such as speed and size by human annotators with the aid of MLLM~\cite{InternVL} to ensure realistic motion simulation. 

\noindent$\bullet$
\textbf{Video Annotations.}~The proposed \datasetname dataset offers thorough annotations, including pose information for both objects and the camera, instance segmentation masks, depth maps, and detailed descriptions of content and motion, broadening its applicability across various fields~\cite{MOSE,MeViS,CharaConsist}.

\subsection{Data Generation Pipeline}\label{sec:data_pipeline}

\noindent$\bullet$
\textbf{Asset Collection and Annotation.}
We collect diverse 3D assets, including environmens and objects. Environment assets span five types:~\emph{ground},~\emph{near ground},~\emph{sky},~\emph{water surface}, and~\emph{underwater}. 
We collect high quality panoramic HDRI images from internet. 
For object assets, we select animated object assets from Objaverse-LVIS~\cite{objaverse}, Objaverse-XL~\cite{objaverse-xl}, Mixamo~\cite{Mixamo}, 
\etc, covering diverse categories. Human annotators filter low-quality assets and verify/correct object properties (\eg, class, habitat, speed, size) queried by InternVL~\cite{InternVL}. They also label motion type and provide description for each animation.

\noindent$\bullet$
\textbf{Object Motion.} To create realistic motions, we design trajectories based on B\'ezier curves in each motion segment, with rotations derived from tangent and normal vectors along the curve. \cref{fig:trajectory_shape} exemplifies motion types. Control points are constrained based on the object's speed.

\begin{figure}
    \centering
    \includegraphics[width=0.999\linewidth]{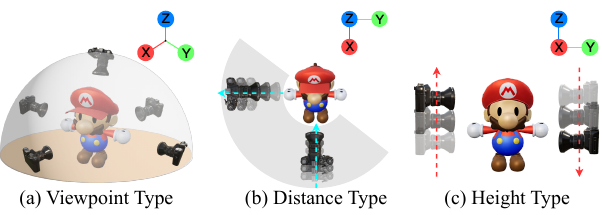}

    \begin{picture}(0,0)
        \scriptsize {
            \put(-112,30){\scalebox{0.8}{\protect \circled{1}}}
            
            \put(-56,30){\scalebox{0.8}{\protect \circled{3}}}

            \put(-54,50){\scalebox{0.8}{\protect \circled{2}}}

            \put(-110,47){\scalebox{0.8}{\protect \circled{4}}}

            \put(-79,78){\scalebox{0.8}{\protect \circled{5}}}
            \put(-31,48){\scalebox{0.8}{\protect \circled{7}}}

            \put(-5,32){\scalebox{0.8}{\protect \circled{6}}}

            \put(39,59){\scalebox{0.8}{\protect \circled{8}}}
            \put(118.5,35){\scalebox{0.8}{\protect \circled{9}}}
        }
    \end{picture}
    \vspace{-6mm}
    \caption{Camera motion types. We decompose camera motion into 3 aspects. (a) \textit{Viewpoint} controls camera orientation when capturing object. \scalebox{0.85}{\protect \circled{1}}-\scalebox{0.85}{\protect \circled{5}} present front/back, left/right and top perspectives. (b) \textit{Distance} and (c) \textit{Height} determine horizontal and vertical distance between camera and object, respectively. \scalebox{0.85}{\protect \circled{6}}-\scalebox{0.85}{\protect \circled{9}} are zoom in/out and up/down, respectively. The ``static'' types are omitted in (b) and (c), which stand for fixed distances.}
    \label{fig:camera_type}
    \vspace{-3.6mm}
\end{figure}

\noindent$\bullet$
\textbf{Camera Motion.}~We decompose camera motion into 3 types:~\emph{viewpoint},~\emph{distance},~and~\emph{height}, see~\cref{fig:camera_type}. (a) \emph{View-point} controls camera’s orientation when capturing objects: {front/back}, {left/right}, and {top}. 
Start and end viewpoints of a motion segment are randomly assigned, with intermediate frames interpolated for smooth transitions, enabling camera’s motion range to encompass all orientations around the object.~(b) \emph{Distance} controls the horizontal distance between the camera and objects: {zoom in/out} and {static}. (c) \emph{Height} controls the vertical distance:~{up/down} and {static}. 
To maintain object visibility without centering it, the camera targets a randomly offset point near the object's centroid.

\begin{table*}[t]
    \centering
    \caption{Comparison of \textbf{\methodname} with other methods. {\methodname excels in controlling 6D poses of objects and camera with diverse motion patterns.}}
    \vspace{-3.6mm}
    \footnotesize
    \setlength{\tabcolsep}{2.6mm}
    \resizebox{1\textwidth}{!}{
    \begin{tabular}{l|cccc}
    \specialrule{.1em}{.05em}{.05em}
        \rowcolor{mygray!50}Methods & Motion Condition & Object Control & Camera Control  & Dataset \\ \hline
        AnimateDiff~\cite{AnimateDiff}&  \xmarkg & \xmarkg & Limited Patterns & Dataset for Specific Motion Pattern \\ 
        
        CameraCtrl~\cite{CameraCtrl}& \textbf{Camera Pose} & \xmarkg & \textbf{Diverse Patterns}   & RealEstate10K~\cite{RealEstate-10K} \\ 
        \rowcolor{mygray2!70}VideoComposer~\cite{VideoComposer} & Image Space Trajectory & Entanglement & Entanglement   & LAION-400M~\cite{LAION} + WebVid~\cite{WebVid-10M} \\
        \rowcolor{mygray2!70}DragNUWA~\cite{DragNUWA} & Image Space Trajectory & Entanglement & Entanglement  & WebVid+ VideoHD~\cite{DragNUWA} \\
        \rowcolor{mygray2!70}3DTrajMaster~\cite{3DTrajMaster} & \textbf{Object Pose}& Limited Patterns & \xmarkg  & \ang{360}-Motion~\cite{3DTrajMaster}  \\
        Direct-a-Video~\cite{Direct-A-Video} &  Image Space Trajectory + Camera Type  & Entanglement & Limited Patterns  & MovieShot~\cite{MovieShot} \\
        MotionCtrl~\cite{MotionCtrl} & Image Space Trajectory + \textbf{Camera Pose}& Entanglement & \textbf{Diverse Patterns}  & RealEstate10K + WebVid  \\ 
        \hline
        \rowcolor{green!8}\textbf{\methodname} (\textbf{ours}) & \textbf{Object Pose + Camera Pose}& \textbf{Diverse Patterns} & \textbf{Diverse Patterns}  & \textbf{\datasetname (ours)}  \\ 
        \specialrule{.1em}{.05em}{.05em}       
     \end{tabular}
     }
     \label{tab:methods}
     \vspace{-3.96mm}
 \end{table*}

\begin{figure}
    \centering
    \includegraphics[width=0.86\linewidth]{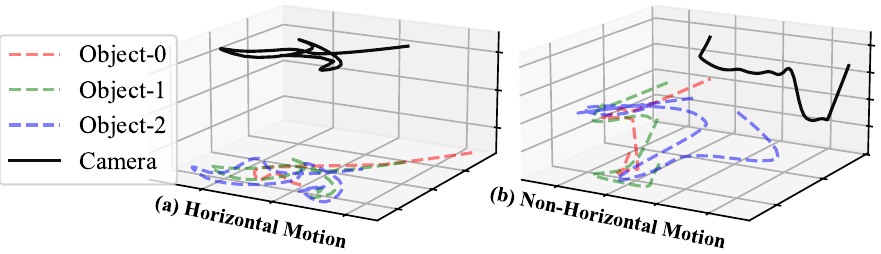}
    \vspace{-2.16mm}
    \caption{{Motion trajectories. The examples of horizontal and non-horizontal motions indicate the variety and complexity of the trajectories generated by our rule-based algorithm.}}
    \label{fig:curves}
    \vspace{-3.16mm}
\end{figure}

\noindent$\bullet$
\textbf{Generation of Multi-Object Scenes.}~Objects from the same environment with comparable sizes and speeds are selected.~The first object’s trajectory is created using the way described earlier, while subsequent trajectories are derived based on the preceding object’s path with a reasonable offset. For camera motion, we randomly track an object in each motion segment, with the trajectory generated using the same method as previously outlined.

\noindent$\bullet$
\textbf{Rendering.} This stage creates synthetic videos based on selected assets and motion types of objects and the camera. We divide a complete motion trajectory into many small segments.~For each segment, we randomly select animations for objects and generate trajectories according to their annotated motion types. Similarly, random combinations of camera motion types are applied across different segments, allowing the generated videos to encompass diverse motion patterns as in real-world scenarios.~These segments are seamlessly combined to form a complex and diverse global trajectory, as shown in \cref{fig:curves}. Finally, the selected assets and poses of the camera and objects are imported into \emph{Unreal Engine}~\cite{unrealengine} to get 3D animation sequence for rendering. \cref{fig:teaser} shows an example trajectory segment.
\section{The Proposed Approach}\label{sec:approach}
Given $N$-length camera poses $\mathcal{C}_{RT}$~=~${RT}_{cam}^{1:N}$, object poses $\mathcal{O}_{RT}$~=~$\{RT_{obj_i}^{1:N}\}_{i=1}^{N_o}$ of $N_o$ objects in the global coordinate system, and content description $\mathbf{C}_p$, we aim to generate the video that reveals correct motion in real world.

As shown in \cref{tab:methods}, most methods cannot independently or jointly control object and camera movements in a 3D-aware manner. For example, AnimateDiff~\cite{AnimateDiff} and CameraCtrl~\cite{CameraCtrl} only support camera control.~Methods~\cite{VideoComposer,DragNUWA,Direct-A-Video,MotionCtrl} using image-space trajectories faces motion entanglement issues and can't control the orientation. Although Direct-a-Video~\cite{Direct-A-Video} introduces several camera types and allows explicit control, it is limited to simple motion patterns. MotionCtrl~\cite{MotionCtrl}, the closest to ours, trains two motion modules separately but lacks comprehensive 6D pose annotations, resulting in suboptimal simultaneous control of camera and object motions. Additionally, it simply applies standard diffusion loss~\cite{Diffusion} when training motion modules, which further hinders its ability to disentangle camera and object motions within a video. \methodname trained on \datasetname introduces \camctrlcompletename (\camctrlname) and \objctrlcompletename (\objctrlname) to address these limitations. \objctrlname receives 6D pose and coarse mask of the object to perceive its spatial location and orientation, achieving a realistic appearance from various viewpoints. The training objectives enable \methodname to disentangle the motion effects of objects and the camera in the video, allowing independent or joint 3D-aware control of camera and object motions.

\noindent$\bullet$~\textbf{{Preliminary}.}~{1)} T2V diffusion models~\cite{AnimateDiff,Imagen-Video,I2VGen-XL,MAGVIT,Make-A-Video,VDM} add Gaussian noise $\epsilon$ to image sequence $\mathbf{z}_0^{1:N}$ in training, resulting in noisy latents $\mathbf{z}_t^{1:N}$ at $t$ time step. Network $\varepsilon_{\theta}$ then is trained to infer the injected noise from current latents.~2) LoRAs~\cite{LoRA,AnimateDiff} are used to learn different content styles.~We apply this to bridge the synthetic-real video domain gap. {3)} Following~\cite{CameraCtrl}, we use {pl\"ucker embedding} to represent camera pose for geometric interpretation.

\begin{figure}[t]
\vspace{-1.6mm}
    \centering
    \includegraphics[width=0.999\linewidth]{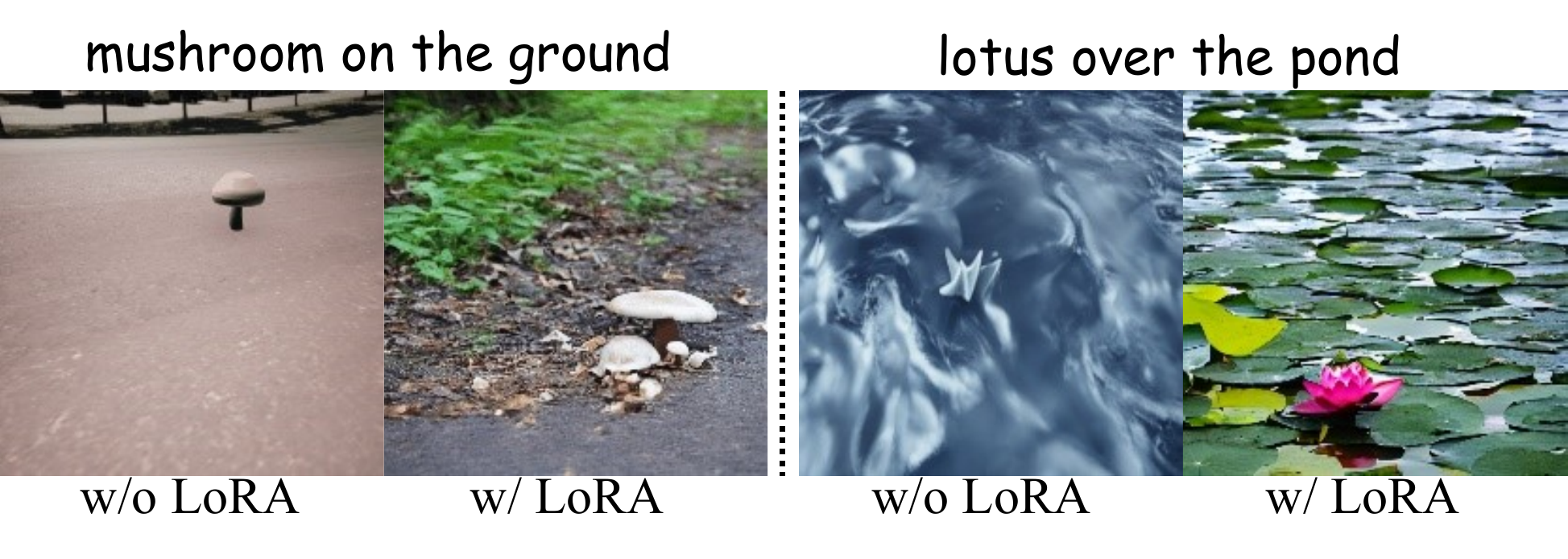}
    \vspace{-8.6mm}
    \caption{Domain LoRA. We sample the first frame of generated videos under without and with Domain LoRA settings.}
    \label{fig:lora}
    \vspace{-3.6mm}
\end{figure}

 \begin{figure*}[t]
    \centering
    \includegraphics[width=0.999\linewidth]{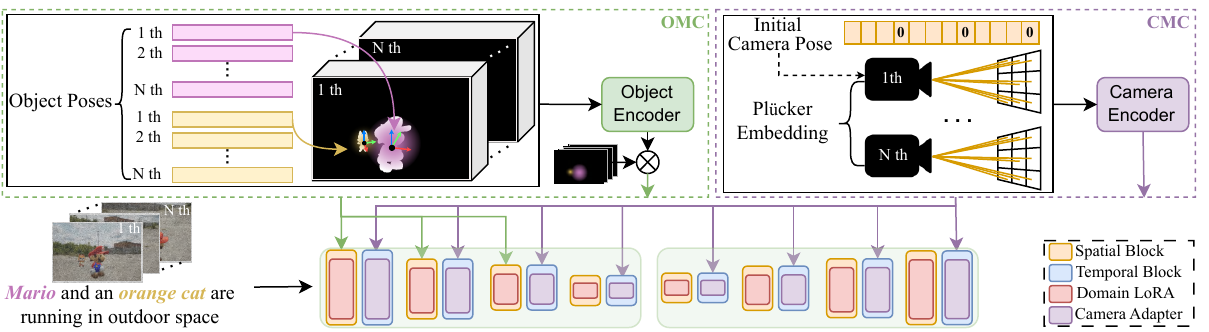}
    \vspace{-6.6mm}
    \caption{The architecture of \methodname. In the first stage, we randomly sample the images from synthetic videos and update the parameters from injected \textit{Domain LoRA}. Next, the modules from \camctrlname are learned. It consists of two parts: \textit{Camera Encoder} and \textit{Camera Adapter}, where the Camera Adapter is introduced into the temporal modules. Finally, we train the \textit{Object Encoder} from \objctrlname. It receives the 6D object pose features, which are repeated in the corresponding object region. We use Gaussian blur kernel centered at the centroid to prevent the need of precise masks. Then, the output is multiplied by the coarse masks to modulate the features in the main branch.}
    \label{fig:network}
    \vspace{-3.6mm}
\end{figure*}

\subsection{Free-Form Motion Control}\label{sec:method}
\cref{fig:network} shows the overall architecture of the proposed \methodname method. We train it in 3 stages.~First, Domain LoRA~\cite{LoRA} are injected into spatial blocks to adapt to rendered content, with temporal modules inactive and images randomly sampled from synthetic data. \cref{fig:lora} shows the effectiveness of this stage in bridging the domain gap.~Then, \camctrlname is trained to learn camera motion, introducing temporal modules and loading LoRA from the previous step. Finally, \objctrlname is trained to decouple object dynamics from camera motion, with other parameters frozen. During inference, LoRA modules are dropped to maintain the quality of base model.

\noindent$\bullet$
\textbf{\camctrlcompletename (\camctrlname).} 
It consists of two parts as shown in \cref{fig:network}. 
Camera Encoder receives pl\"ucker embeddings, where the initial camera pose (translation values are set to 0) and the relative camera poses are used for the first and subsequent frames respectively. The initial pose helps to determine the perspective at the start time. Then, the outputs are processed by Camera Adapter to modulate the features in temporal blocks. Due to the dynamic of the background being only affected by camera motion, \textit{camera loss} $L_{cam}$ is applied in this stage:
\vspace{-0.6mm}\begin{equation}\vspace{-0.6mm}
\resizebox{0.432\textwidth}{!}{$
\begin{aligned}\label{eq:cam-ddpm-loss}
L_{cam}=E_{\mathbf{z}_0^{1:N},t,\epsilon,\mathbf{C}_p,\mathcal{C}_{RT}}&[\mathcal{M}_{bg}\left\|\varepsilon_{\theta,\theta_c}\left(\mathbf{z}^{1:N}_t, t, \mathbf{C}_p,\mathcal{C}_{RT}\right)-\epsilon\right\|^2 \\
&+ \lambda_c \left\|\varepsilon_{\theta,\theta_c}\left(\mathbf{z}^{1:N}_t, t, \mathbf{C}_p,\mathcal{C}_{RT}\right)-\epsilon\right\|^2 ],
\end{aligned}
$
}
\end{equation}
where $\theta_c$ is the parameters of \camctrlname. $\mathcal{M}_{bg}$ is background mask and $\lambda_c$ is the weighting factor. $L_{cam}$ makes camera motion more accurate by concentrating on the background.

\noindent$\bullet$
\textbf{\objctrlcompletename (\objctrlname).} The Object Encoder of \objctrlname receives 6D object pose information to adjust the features in spatial modules from several downsample blocks~\cite{StableDiffusion,MotionCtrl}. Specifically, the poses relative to the camera in each frame are duplicated within the respective object region while the others are set to 0. Then, the pose features concatenated with the foreground mask are fed to \objctrlname. In this manner, the poses from different objects can be aggregated in a single input. Besides, we leverage the Gaussian blur kernel centered at the object centroid to avoid users offering precise masks. Then, the outputs from \objctrlname are multiplied by the coarse masks and added to the spatial features in the main branch, preventing to impair the background content. During inference, the size of the kernel can be approximated based on the object's size (specified by the user) and its distance from the camera. \textit{object loss} $L_{obj}$ is applied to make \objctrlname focus on the object region:
\vspace{-1.6mm}\begin{equation}\vspace{-1mm}
\resizebox{0.332\textwidth}{!}{$
\begin{aligned}\label{eq:obj-ddpm-loss}
&L_{obj}=E_{\mathbf{z}_0^{1:N},t,\epsilon,\mathbf{C}_p,\mathcal{C}_{RT},\mathcal{O}_{RT}}[ \\
&\mathcal{M}_{fg}\left\|\varepsilon_{\theta,\theta_c,\theta_o}\left(\mathbf{z}^{1:N}_t, t, \mathbf{C}_p,\mathcal{C}_{RT},\mathcal{O}_{RT}\right)-\epsilon\right\|^2  \\
&+\lambda_o \left\|\varepsilon_{\theta,\theta_c,\theta_o}\left(\mathbf{z}^{1:N}_t, t, \mathbf{C}_p,\mathcal{C}_{RT},\mathcal{O}_{RT}\right)-\epsilon\right\|^2 ],
\end{aligned}
$}
\end{equation}
where $\theta_o$ indicates the parameters from \objctrlname. $\mathcal{M}_{fg}$ is foreground mask and $\lambda_o$ is the weighting factor.~$L_{obj}$ improves the appearance quality of dynamic objects.

\section{Experiments}
\label{sec:experiments}

\noindent$\bullet$
\textbf{Implementation Details.} The proposed \methodname is based on AnimateDiff V3~\cite{AnimateDiff}, trained with 16-length $256 \times 384$ videos, Adam optimizer with a learning rate of $1e^{-4}$. Domain Adapter is trained with 8K iterations in a batch size of 128. \camctrlname and  \objctrlname are trained with 50K iterations with a batch size of 8. $\lambda_c$ and $\lambda_o$ are set to 0.6 and 0.3. In experiments, we find that 26K data samples suffice for our method to learn 3D-aware motion control.

\noindent$\bullet$
\textbf{Evaluation Metrics.} Following \cite{AnimateDiff,MotionCtrl}, we use FID~\cite{FID} to evaluate visual quality, FVD~\cite{FVD} for temporal coherence, and CLIPSIM~\cite{CLIP} to measure semantic similarity with text. For camera motion, we follow~\cite{CameraCtrl} to use CamTransErr and CamRotErr. 
For object motion, ObjTransErr and ObjRotErr are introduced. We first use depth estimation model~\cite{DepthAnythingV2} to obtain the depth at object centroid and determine its global position based on camera pose, then fit a trajectory curve to find tangent and normal vectors at each time step, allowing for rotation derivation. Given scale information, we apply appropriate scaling to the translation error calculation.

\begin{figure*}[t]
    \centering
    \includegraphics[width=0.999\linewidth]{indep_comparison_compressed.pdf}
    \vspace{-7.6mm}
    \caption{Independent controls over camera and object motions. Results in (a) reveal that all methods~\cite{CameraCtrl,MotionCtrl} effectively reflect the camera conditions. For object motion, the compared methods~\cite{MotionCtrl,Direct-A-Video} fail to maintain a stationary camera as shown in green boxes from (b) (\eg, movement of flower in row 5). Furthermore, they also present low fidelity of object orientation (3D axes in conditions).}
    \label{fig:indep_comparison}
    \vspace{-3.06mm}
\end{figure*}

\subsection{Comparisons with State-of-the-Art Methods} 

 We first compare independent controls over camera motion and object motion with previous methods \cite{MotionCtrl,CameraCtrl}. Then, we demonstrate \methodname’s superior performance in simultaneous control. Finally, we showcase additional examples across different scenes to validate the effectiveness of \datasetname and \methodname. For fairness, we compare with U-Net based methods. More examples are in supplementary.

\noindent$\bullet$
\textbf{Independent~Control~of~Camera~Motion.} MotionCtrl \cite{MotionCtrl} and CameraCtrl \cite{CameraCtrl} are selected for this comparison as they accept explicit camera information. In \cref{fig:indep_comparison}(a), we simulate two camera motions and scale the translation to fit the input range required by these methods.~\methodname and the compared methods effectively reflect the input conditions. The CamTransErr~\&~CamRotErr in \cref{tab:performance_metrics} also show that \methodname achieves comparable results in controlling camera.

\begin{table}[t]
\centering
\caption{Quantitative comparison of our proposed method \methodname with AnimateDiff \cite{AnimateDiff}, CameraCtrl \cite{CameraCtrl}, and MotionCtrl \cite{MotionCtrl}.}
\label{tab:performance_metrics}
\vspace{-3mm}
\footnotesize
\setlength{\tabcolsep}{0.916mm}
\begin{tabular}{l|c|c|c|c}
\specialrule{.1em}{.05em}{.05em} 
\rowcolor{mygray!50}Method & AnimateDiff & CameraCtrl & MotionCtrl & \textbf{\methodname (ours)} \\
\hline\hline
\textbf{FID} $\downarrow$ & 149.61 & 137.96 & \textbf{125.52} & \underline{133.42} \\
\rowcolor{mygray2!36}\textbf{FVD} $\downarrow$ & 868.97 & \textbf{805.25} & 952.31 & \underline{846.51} \\
\textbf{CLIPSIM} $\uparrow$ & \ \ 29.33 & \ \  29.21 & \ \ 26.83 & \ \ \textbf{31.01} \\
\rowcolor{mygray2!36}\textbf{CamTransErr} $\downarrow$& - & \ \  18.16 & \ \  \textbf{17.84} & \ \ \underline{18.12} \\
\textbf{CamRotErr} $\downarrow$ & -& \ \  \ \ \textbf{0.94} & \ \ \ \ 1.11 & \ \ \ \ \underline{1.03} \\
\rowcolor{mygray2!36}\textbf{ObjTransErr} $\downarrow$ & -& - & \ \  80.66 & \ \ \textbf{42.25} \\
\textbf{ObjRotErr} $\downarrow$ & -& - & \ \ \ \  1.77 &\ \ \ \ \textbf{0.96} \\
\specialrule{.1em}{.05em}{.05em} 
\end{tabular}
\vspace{-6mm}
\end{table}

\begin{figure}[t]
    \centering
    \includegraphics[width=0.999\linewidth]{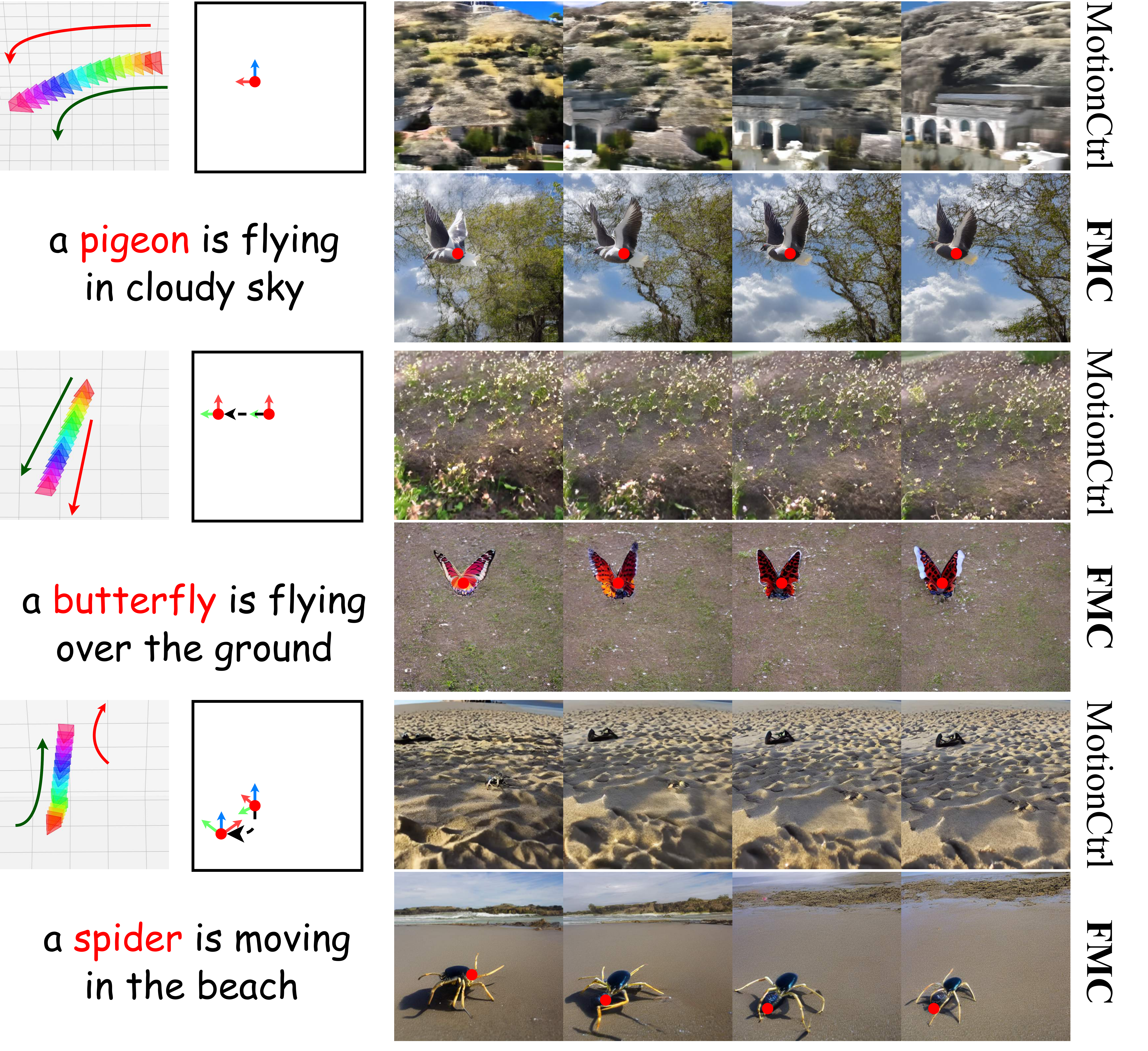}
    \vspace{-7.8mm}
    \caption{Simultaneous control over camera and object motions. MotionCtrl \cite{MotionCtrl} struggles to generate realistic object dynamics, causing objects to disappear from view, whereas our \methodname achieves high-quality simultaneous control.}
    \label{fig:simul_comparison}
    \vspace{-5.6mm}
\end{figure}

\noindent$\bullet$
\textbf{Independent~Control~of~Object~Motion.} For object control, we compare with MotionCtrl~\cite{MotionCtrl} and Direct-a-Video \cite{Direct-A-Video}.~For these methods, we project the global trajectory into image space using camera and object pose information. As shown in \cref{fig:indep_comparison}(b), 
the compared methods fail to maintain a stationary camera (showing dynamic movement in the background),
indicating that image space trajectories entangle the object and camera motions. For example, the 2nd example of \cref{fig:indep_comparison}(b), Direct-a-Video~\cite{Direct-A-Video} shows the change of flower location caused by dynamic camera. Our method effectively alleviates this issue by incorporating static camera poses as constraints. {Furthermore, since \objctrlname receives 6D pose of objects, \methodname can achieve high fidelity of object orientation with input condition.}

\begin{table}[t]
\vspace{-2mm}
\centering
\caption{User study in quality, text similarity, and motion fidelity.}
\label{tab:user_study}
\vspace{-3mm}
\footnotesize
\setlength{\tabcolsep}{0.96mm}
\resizebox{0.476\textwidth}{!}{
\begin{tabular}{l|c|c|c}
\specialrule{.1em}{.05em}{.05em} 
\rowcolor{mygray!50}Method & CameraCtrl~\cite{CameraCtrl} & MotionCtrl~\cite{MotionCtrl}  & \textbf{\methodname (ours)} \\
\hline\hline
\textbf{Quality Score} & 0.88 & \underline{0.89} & \textbf{0.91} \\
\rowcolor{mygray2!36}\textbf{Text Similarity Score} & \underline{0.84} & 0.81 & \textbf{0.95} \\
\textbf{Camera Motion Score} & \textbf{0.95} & \underline{0.93} & \textbf{0.95} \\
\rowcolor{mygray2!36}\textbf{Object Motion Score}& - & 0.53 &  \textbf{0.98} \\
\specialrule{.1em}{.05em}{.05em} 
\end{tabular}
}
\vspace{-3mm}
\end{table}

\begin{table}[t]
\centering
\caption{Quantitative results in ablation study.}
\label{tab:ablation_quantitative}
\vspace{-3mm}
\footnotesize
\setlength{\tabcolsep}{0.6mm}
\resizebox{0.47\textwidth}{!}{
\begin{tabular}{l|c|c|c|c}
\specialrule{.1em}{.05em}{.05em} 
\rowcolor{mygray!50}Metrics & CamTransErr & CamRotErr & ObjTransErr & ObjRotErr \\
\hline\hline
MotionCtrl (\textit{w/o} $\mathcal{C}_{RT}$)  & 18.24 & 1.08 & 78.82 & 1.65\\
\rowcolor{mygray2!36} MotionCtrl (\textit{w/} $\mathcal{C}_{RT}$) & 18.24 & 1.08 & 55.33 & 1.26 \\
\textbf{FMC} (\textit{w/o} $L_{cam}$) & 20.35 &1.19 & - & - \\
\rowcolor{mygray2!36}\textbf{FMC} (\textit{w/o} $L_{obj}$)  & \textbf{18.12}&  \textbf{1.03} & 46.62 & 1.15 \\
\textbf{FMC}  & \textbf{18.12}&  \textbf{1.03} &\textbf{42.25} & \textbf{0.96} \\
\specialrule{.1em}{.05em}{.05em} 
\end{tabular}
}
\vspace{-7mm}
\end{table}

\vspace{-.5mm}
\noindent$\bullet$
\textbf{Simultaneous Control of Camera and Object Motions.} We explore the combination of both camera and object control signals. Since Direct-a-Video~\cite{Direct-A-Video} supports only basic camera motion, we choose MotionCtrl~\cite{MotionCtrl} as the comparison method. We randomly simulate movements for both the object and the camera, resulting in varied trajectories. As shown in \cref{fig:simul_comparison}, videos generated by \methodname more faithfully align with the specified conditions. While MotionCtrl captures the camera’s motion, it struggles to generate realistic object dynamics. 
These results demonstrate the effectiveness of \methodname in achieving simultaneous control of camera and object motions. The object error metrics in \cref{tab:performance_metrics} show that our method achieves better results in object motion control. Additionally, \methodname achieves higher scores in user study as shown in \cref{tab:user_study}, outperforming previous methods in quality and motion fidelity.

\begin{figure}[t]
    \centering
    \includegraphics[width=0.999\linewidth]{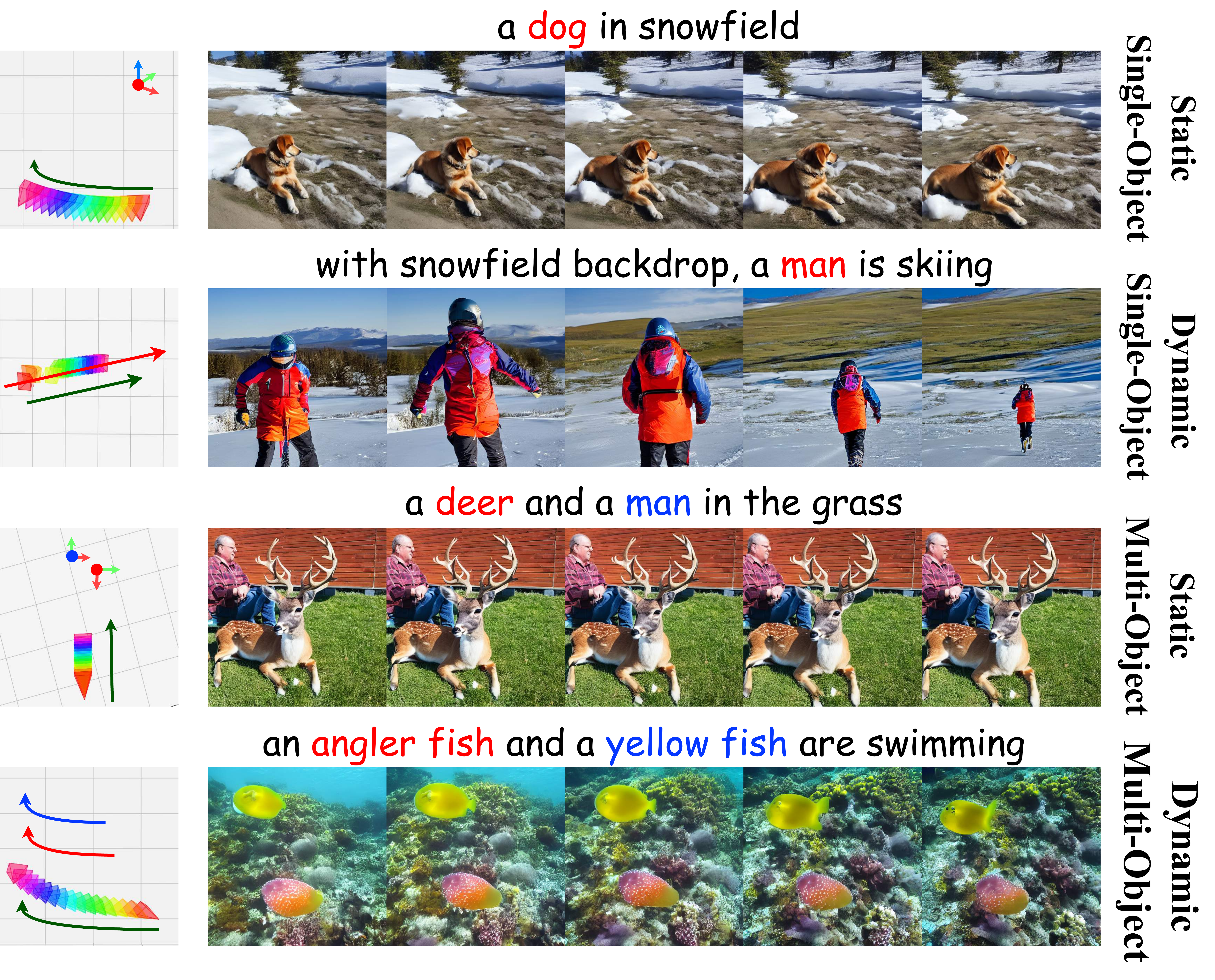}
    
    \vspace{-3.96mm}
    \caption{Simultaneous control results of \methodname in different cases. The complicated case in row 2 shows that our method learns complex shot, where the camera first captures a skier from the front and then follows him from behind. }
    \label{fig:simul}
    \vspace{-2.6mm}
\end{figure}

\cref{fig:simul} shows video generation results across 4 different cases: \textit{static single-object}, \textit{dynamic single-object}, \textit{static multi-object}, and \textit{dynamic multi-object}. Thanks to the diversity of motion patterns in \datasetname, \methodname effectively learns a range of diverse, advanced, and complex shots. In the {2nd} row of \cref{fig:simul}, for example, the camera initially captures the person from the front and then follows from behind. The last two rows demonstrate the performance in multi-object scenarios, where the relative motion between objects and the camera conforms closely to the input conditions.

\subsection{Ablation Study}

\begin{figure}
    \centering
    \includegraphics[width=0.999\linewidth]{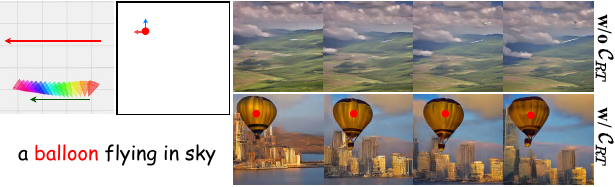}
    \vspace{-7.6mm}
    \caption{Simultaneous control results of MotionCtrl~\cite{MotionCtrl} trained on \datasetname without and with camera pose during training.}
    \label{fig:ablation_1}
    \vspace{-4.6mm}
\end{figure}

\noindent$\bullet$ 
\textbf{\datasetname Dataset.}~To validate the effectiveness of complete camera and object pose annotations in \datasetname and evaluate the dataset generalization, we train MotionCtrl~\cite{MotionCtrl} on it, adapting the annotations to its input format. We first train the camera module, then optimize the object module in two ways: without camera poses, as in \cite{MotionCtrl}, and with camera poses, as our \methodname. As shown in \cref{fig:ablation_1}, incorporating known camera poses when optimizing object module allows the object to follow the input trajectory more accurately, reducing the risk of it leaving the field of view. Object motion errors in \cref{tab:ablation_quantitative} further highlight the benefits of using complete camera and object pose annotations.

\begin{figure}
    \centering
    \includegraphics[width=0.999\linewidth]{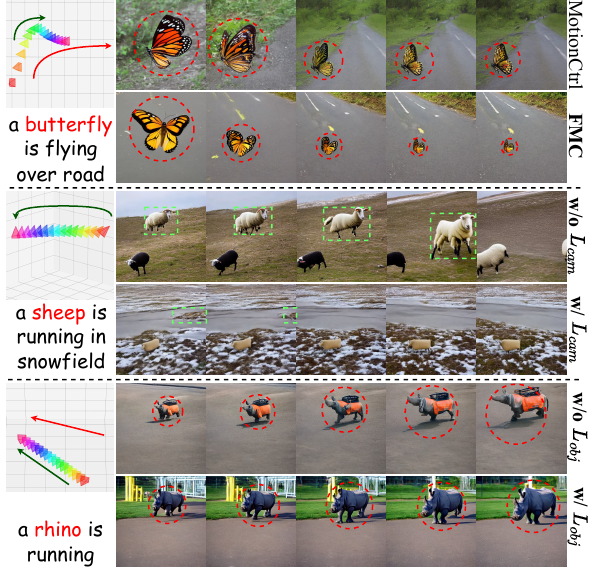}
    \vspace{-7.6mm}
    \caption{{Results of different settings in the ablation study. The first row is MotionCtrl~\cite{MotionCtrl} trained on \datasetname.}}
    \label{fig:ablation_2}
    \vspace{-4.6mm}
\end{figure}

\noindent$\bullet$
\textbf{OMC.}~{As shown in the first 2 rows of \cref{fig:ablation_2} and object motion errors in \cref{tab:ablation_quantitative}, \methodname outperforms MotionCtrl~\cite{MotionCtrl} trained on \datasetname in motion accuracy. This improvement is brought by \objctrlname's ability to process 6D poses, generating more realistic object appearances based on orientation and object's distance from the camera, reflected by the size of coarse masks.~\cite{MotionCtrl}'s object motion control module can only handle 2D image-space trajectories without pose and distance information, limiting alignment accuracy with input.}

\noindent$\bullet$
\textbf{Training Objectives.} We conduct two experiments to assess the impact of $L_{cam}$ in \cref{eq:cam-ddpm-loss} and $L_{obj}$ in \cref{eq:obj-ddpm-loss}. First, we train \camctrlname using only the standard diffusion loss~\cite{Diffusion,SDE}. As shown in the 3rd row of \cref{fig:ablation_2}, model without $L_{cam}$ tends to shift foreground objects to achieve similar relative motion, which does not accurately match the input pose. The camera motion error in \cref{tab:ablation_quantitative} underscores the effectiveness of $L_{cam}$. Besides, training \objctrlname without $L_{obj}$ leads to undesired object appearances, as shown in the 5th row of \cref{fig:ablation_2}, with object motion error in \cref{tab:ablation_quantitative} confirming the benefit of $L_{obj}$.

\noindent$\bullet$
\textbf{Adaptability with Different T2I Personalized models.} As shown in \cref{fig:personalized}, \methodname can be adapted to various personalized backbones~\cite{Toonyou,RealisticVision,StableDiffusion}, showing that our proposed dataset \datasetname and corresponding training strategy do not impair the model's original generative capabilities.

\vspace{-0.8mm}
\section{Conclusion}
\label{sec:conclusion}
\vspace{-1mm}

This work introduces \datasetname, a dataset with comprehensive 6D pose information and diverse assets, offering both standard and complex shots that are difficult to capture in real life, with trajectories resembling real-world scenarios. With \datasetname, the proposed method \methodname enables independent or joint 3D-aware control of object and camera motions within a single video. Experimental results demonstrate the effectiveness of both \datasetname dataset and \methodname.

\noindent\textbf{Limitations.}~Our method's ability to control complex motions of multiple objects remains limited. Better metrics are needed to more accurately evaluate object motion. In the future, additional input modalities, \eg, images, are desired to customize motion videos for reference subjects.

\clearpage
\footnotesize{\paragraph{Acknowledgement.} This project was supported by the National Natural Science Foundation of China (NSFC) under Grant No.~62472104. This work was supported by Damo Academy through Damo Academy Innovative Research Program. Dr Tao's research is partially supported by NTU RSR and Start Up Grants. This work was in part supported by National Natural Science Foundation of China (Grant No. 62276067).}

{
    \small
    \bibliographystyle{ieeenat_fullname}
    \bibliography{main}
}

\end{document}